\documentclass[a4paper, 10 pt, conference]{ieeeconf} % Use this line for a4 paper

\IEEEoverridecommandlockouts % The preceding line is only needed to identify funding in the first footnote. If that is unneeded, please comment it out.

\usepackage{cite}
\usepackage{amsmath,amssymb,amsfonts}
\usepackage{algorithmic}
\usepackage{graphicx}
\usepackage{textcomp}
\usepackage{xcolor}

\usepackage{caption}
\usepackage{subcaption}

\usepackage{algorithm}

\usepackage{array}

\usepackage{hyperref}

\def\BibTeX{{\rm B\kern-.05em{\sc i\kern-.025em b}\kern-.08em
    T\kern-.1667em\lower.7ex\hbox{E}\kern-.125emX}}

\begin{document}

\title{Real-time Autonomous Glider Navigation Software}

\author{Ruochu Yang, Mengxue Hou, Chad Lembke, Catherine Edwards, and Fumin Zhang
\thanks{
The research work is supported by ONR grants  N00014-19-1-2556 and N00014-19-1-2266;  AFOSR grant FA9550-19-1-0283; NSF grants GCR-1934836,  CNS-2016582 and ITE-2137798; and NOAA grant NA16NOS0120028.
}
\thanks{Ruochu Yang and Fumin Zhang are with the School of Electrical and Computer Engineering, Georgia Institute of Technology, Atlanta, USA. Mengxue Hou is with the College of Engineering, Purdue university, West Lafayette, USA. Chad Lembke is with the College of Marine Science, University of South Florida, St.Petersburg, USA. Catherine Edwards is with the Skidaway Institute of Oceanography, University of Georgia, Savannah, USA}
}

\maketitle

\begin{abstract}

Underwater gliders are widely utilized for ocean sampling, surveillance, and other  various oceanic applications. In the context of  complex ocean environments,  gliders may yield poor navigation performance due to strong ocean currents, thus requiring substantial human effort during the manual piloting process. To enhance navigation accuracy,  we developed a real-time autonomous glider navigation software, named GENIoS\_Python,  which generates waypoints based on flow predictions to assist human piloting. The software is designed to closely check glider status, provide customizable experiment settings, utilize lightweight computing resources, offer stably communicate with dockservers, robustly run for extended operation time, and quantitatively compare flow estimates, which add to its value as an autonomous tool for underwater glider navigation.

\end{abstract}

\begin{keywords}
autonomous underwater glider navigation, ocean flow prediction
\end{keywords}

\section{Introduction}
\label{intro}
Underwater gliders have been widely used in ocean fields for ocean sampling, surveillance, and many other applications  \cite{doi:10.1080/00207170701222947, 4476150, hou2020, nicholson2008present, schofield2007slocum}. Usually, the glider is navigated through waypoints manually generated by glider pilots. Nonetheless, manual navigation performance might not be optimal, as the manual waypoints fail to consider the strong and erratic ocean flow. Worse, the glider can drift to an unexpected area where localization and rescue are extremely tough. Additionally, manual glider navigation requires tremendous human labor. In order to circumvent trajectory deviation or abort issues, the glider needs to be closely monitored during the whole mission.  Considering the long duration of mission, up to ten glider pilots may take shifts  each day, including at midnight. Therefore, there is a pressing need for the development of  a glider navigation software that can generate real-time waypoints based on flow predictions to improve navigation accuracy. The software should also achieve autonomous functionality to significantly reduce the heavy human labor of manual piloting.

Generally, underwater gliders use the global positioning system (GPS) to localize themselves on the ocean surface \cite{iet:/content/conferences/10.1049/cp_19940567}. However, considering that GPS signals can not transmit in the deep ocean, the next waypoint of glider can only be decided at surfacing events with GPS update, which makes it hard to perform underwater glider navigation. Controlled by buoyancy and mass, low-speed gliders are strongly affected by ocean currents, so this type of navigation may lead to huge trajectory deviation as the mission is going on. Some gliders try to estimate the flow between the last and current surfacing events and utilize it to promote  navigation accuracy \cite{soton54740}. This navigation method based on glider-derived flow estimate may work well in a calm water environment, but in the deep ocean with intense temporal and spatial flow changes, the glider-derived flow estimate may differ from the actual flow a lot, causing poor underwater navigation. So far, it is obvious that one main limitation of glider navigation is insufficient knowledge of the ocean environment, especially ocean currents. Recently in the robotics literature, there has been active research on navigation of underwater gliders based on flow predictions \cite{Witt2009GoWT, soulignac2010feasible, pathplanningforgliders, pereira2013risk, szwaykowska2013trend }. Some of them incorporated flow data from ocean models to increase prediction accuracy \cite{reikard2011forecasting, ADCIRC, bleck2002oceanic, shchepetkin2005regional, haidvogel2008ocean, frolov2012improved, kalnay2003atmospheric, lermusiaux2006uncertainty}. However, none of the authors have practically implemented a navigation system for real underwater gliders or experimentally validated their navigation algorithms in real-world scenarios.

The main contribution of this paper is to develop a real-time autonomous glider navigation software based on the Glider-Environment Network Information System (GENIoS) \cite{chang2015real}. The  software can generate waypoints in real time based on flow predictions to better navigate gliders in the field of highly variable ocean flow. The operation is meant to be autonomous in order to relieve a part of heavy labor of glider pilots. Additionally, the software possesses the following features \begin{itemize}
  \item continuous monitoring of glider status.
  
  \item customizable settings of gliders and deployments.
  
  \item lightweight requirements of computing resources.

  \item stable communication with dockservers.

  \item robust operation for extended time.

  \item quantitative comparison of flow predictions.
  
\end{itemize}
Herein, the software is officially named as \textbf{GENIoS\_Python}. Please check the project website \url{https://sites.google.com/view/geniospython}.  We will release our code soon.

\section{System Design}
\begin{figure}[ht]
        \centerline{\includegraphics[width=0.47\textwidth]{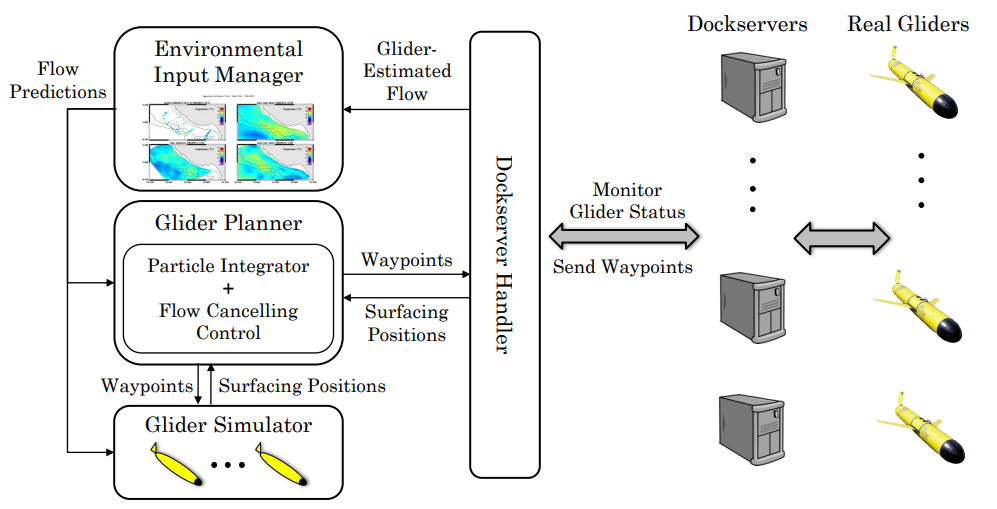}}
        \caption{System design of GENIoS\_Python}
        \label{system design}
    \end{figure}

The prototype of GENIoS\_Python is GCCS (Glider Coordinated Control System) \cite{4476150}, which generates waypoints for a fleet of gliders so that gliders can sample optimally distributed measurements, improving collective survey performance. The navigation performance of GCCS was verified in the Adaptive Sampling and Prediction (ASAP) project\cite{leonard2010coordinated}  in Monterey Bay, California (2004 - 2006). To deal with highly variable flow, we extended GCCS into GENIoS  \cite{chang2015real}  by incorporating real-time ocean flow data and glider surfacing data, which supported the Long Bay missions in South Carolina (2012 - 2013). Building off the success of GENIoS, here comes the newest software GENIoS\_Python. As shown in Fig.~\ref{system design}, GENIoS\_Python consists of four main  modules: glider simulator (gsim), glider planner (gplan), environmental input manager, and dockserver handler.  

The gsim module supports two running modes: simulated and remote. In the simulated mode, the module simulates the glider's underwater trajectory and surfacing positions according to the glider's 3D kinematics model.  Rather than wait a long time (e.g., four hours) for gliders to surface in the real-time deployment, the surfacing interval in the simulated mode can be adjusted as short as possible (e.g., one minute). Therefore, it is effective to verify any underwater navigation algorithm, which may not be practical for validation in real life, with customized glider and deployment settings in an extremely fast manner. In the remote mode, the module simply obtains the glider surfacing  positions through the dockserver handler module in real time.

The gplan module is composed of two classes: path tracking and path planning. The path tracking class provides two tracking algorithms: virtual mooring and line control. The virtual mooring algorithm keeps the glider moving towards one single target position, while the line control algorithm keeps the glider moving back and forth between multiple target positions. The core of the tracking algorithms is a flow-cancelling controller, which computes the desired glider heading to cancel out the predicted flow at the current glider location. Then, by assuming the glider as a Newtonian particle \cite{4476150} and integrating its motion with the heading control, the gplan module can predict the glider trajectory in the next 12 or 24 hours under the influence of flow. Based on the predicted trajectory, the waypoints are computed for the next 12 or 24 hours as well. By following these waypoints, the glider can track the expected trajectory. The path planning class offers a modified  $A^\ast$  algorithm \cite{hart1968formal} that considers the entire flow field, as opposed to only one single location in the path tracking class. The  algorithm plans an optimal path for the glider to avoid the strong flow field in advance, thus consuming much more time or energy. The planned path can be viewed as a high-level path for the path tracking algorithm to follow.

The environmental input manager module generates flow predictions  for path tracking class  to compute waypoints or for path planning class to compute the  optimal path. The module supports two modes of predicting flow: simulated and remote. In the simulated mode, the module incorporates the ADCIRC ocean flow model to enable the simulation of glider movement in any flow field. In the remote mode, the module utilizes predictive flow data from the oceanic data model Advanced Circulation (ADCIRC) \cite{ADCIRC} in real time.  A hybrid model GliADCIRC,  based on the ADCIRC model, is developed to produce more accurate flow predictions by combining glider-estimated flow with ADCIRC predictive flow.

The dockserver handler module interfaces with an onshore computer (dockserver) to obtain the latest glider surfacing data and send waypoints to gliders in real time. When the glider surfaces,  it communicates with the dockserver through the SFMC (Slocum Fleet Mission Control) terminal that records the glider-transmitted data as log files. The module continuously monitors the  log files to check the latest surfacing event and grab the navigation information for the gplan module to compute waypoints. Relied on fast performance of Python SSH and SFTP packages, the module can check the SFMC terminal every 10 seconds, thus capturing every single surfacing event, even though the glider surfaces for an extremely short time or uses mixed surfacing modes. Then, the module sends the computed waypoints to the dockserver. It takes only 30 seconds out of 10-15 minute surfacing interval to accomplish the entire process of checking surfacing information and transmitting waypoints. Moreover, the module utilizes Linux grep packages to parse glider log files and handle glider data without human interaction, enabling autonomous glider status check.

\section{Experiments}

The performance of GENIoS\_Python was validated in both simulated and real experiments. All the experiments were implemented in a Lenovo ThinkCentre Desktop with Ubuntu 16.04 LTS, Intel Core i7-6700 CPU @ 3.40GHz × 8, 32 GB Memory, and Intel HD Graphics 530 (Skylake GT2). 

\subsection{Simulated Experiments}
The simulated experiments were implemented in the simulated mode of gsim module. The simulated glider setting was adjusted from the Slocum G3 glider Franklin of Skidaway Institute of Oceanography, University of Georgia. The deployment was set in Gray's Reef near Savannah, Georgia, United States. The ocean flow model was set as ADCIRC. We tested two tracking algorithms of the gplan module: virtual mooring and line control.

\subsubsection{Virtual Mooring}
In the virtual mooring mode, both the starting point and the target point were pre-set for the glider Franklin. As seen from the video \url{https://www.youtube.com/watch?v=5KFjQUZV7rU}, GENIoS\_Python successfully navigated the glider Franklin to move directly towards the red target point.

\subsubsection{Line Control}
In the line control mode, two target points were given and the starting point was set as one of them. As seen from the video \url{https://www.youtube.com/watch?v=ObHONQhCt04}, GENIoS\_Python successfully navigated the glider to move back and forth between two red target points.

\subsection{Real Experiment}

The real experiment was carried out on  March in the remote mode of gsim module. The Slocum G1 glider USF-SAM was provided by the glider team of University of South Florida. The deployment field was in Gray's Reef near Savannah, Georgia, United States. The ocean flow model was set as GliADCIRC. The tracking algorithm was set as virtual mooring. GENIoS\_Python was set to check the glider status (header) every ten seconds. During the whole experiment, GENIoS\_Python successfully detected every surfacing event and transmitted the waypoints (goto file) to the dockserver. The experiment was divided into two journeys with two different target points, respectively.

\subsubsection{Journey 1}
\begin{figure}[ht]
        \centerline{\includegraphics[width=0.47\textwidth]{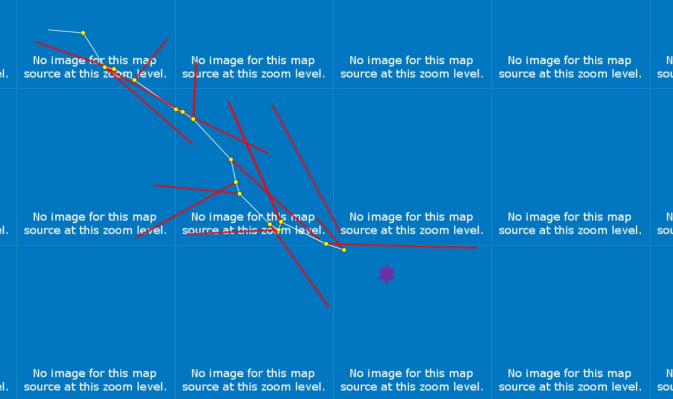}}
        \caption{Real-time trajectory of glider USF-SAM in SFMC terminal. The red lines represent the depth-averaged ocean current. The yellow dots represent the real-time glider  positions. The purple star represents the target point.}
        \label{journey 1}
    \end{figure}

The first journey happened from March 2nd, 2023 to March 4th, 2023 UTC. The target point was 3118.0N, -8008.0E. As shown in Fig.~\ref{journey 1}, the real-time trajectory of the glider USF-SAM was directly approaching the target point under the navigation of GENIoS\_Python. In terms of the full experimental result, it can be seen from the video \url{https://www.youtube.com/watch?v=AcpguPaL5Ik} that GENIoS\_Python successfully navigated the glider USF-SAM to achieve the red target point without twists and turns, saving considerable glider power and mission time.

\subsubsection{Journey 2}
\begin{figure}[ht]
        \centerline{\includegraphics[width=0.47\textwidth]{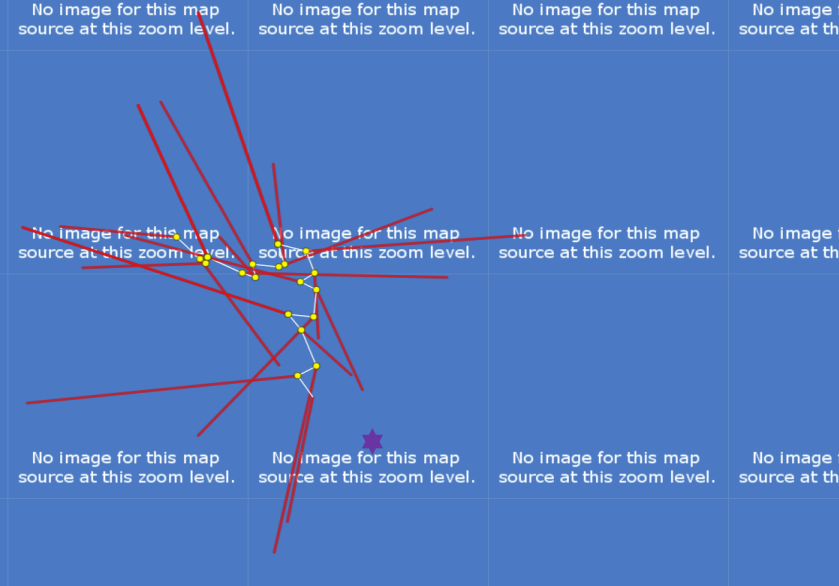}}
        \caption{Real-time trajectory of glider USF-SAM in SFMC terminal. The red lines represent the depth-averaged ocean current. The yellow dots represent the real-time glider  positions. The purple star represents the target point.}
        \label{journey 2}
    \end{figure}

The second journey happened from March 4th , 2023 to March 6th, 2023 UTC. The target point was 3110.0N, -8000.00E. It is obvious from Fig.~\ref{journey 2} that the glider USF-SAM was approaching the target point under the navigation of GENIoS\_Python. In terms of the full experimental result, it can be seen from the video \url{https://www.youtube.com/watch?v=R75oJJ1-U7Q} that GENIoS\_Python successfully navigated the glider USF-SAM to move towards the red target point almost in a straight line.

\begin{figure}[ht]
        \centerline{\includegraphics[width=0.47\textwidth, height=0.5\textwidth]{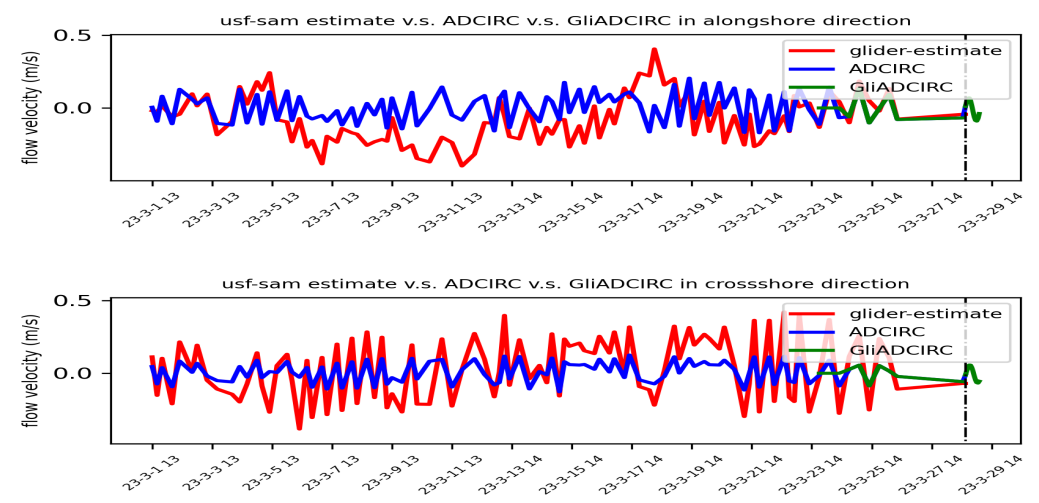}}
        \caption{Comparison of  glider-estimated flow, ADCIRC flow predictions and GliADCIRC flow predictions. The red line represents the glider-estimated flow. The blue line represents ADCIRC flow predictions. The green line represents GliADCIRC flow predictions. The green line behind the black dashed line of timestamp represents GliADCIRC flow predictions for  the upcoming 12 hours following the end of deployment.}
        \label{flow}
    \end{figure}

In order to assist glider pilots with flow analysis, GENIoS\_Python provides the real-time comparison of glider-estimated flow, ADCIRC flow predictions and GliADCIRC flow predictions. If both the glider-estimated flow and flow predictions are larger than $0.3m/s$, it is most likely that the glider has stepped into the strong flow field where the flow speed may exceed the maximum glider speed, causing significant trajectory deviation. The flow comparison of the whole USF-SAM deployment was shown in Fig.~\ref{flow}. For better visualization, the flow is divided into alongshore component and crossshore component. Considering that the gplan module relies on the GliADCIRC model to compute waypoints for the next 12 or 24 hours in real life, the flow compassion function offers GliADCIRC flow predictions for  the upcoming 12 hours following the end of deployment (green line behind the black dashed line of timestamp).

\section{Conclusion}

In summary, underwater gliders play an important role in diverse oceanic applications. Nevertheless, their navigation performance can be hindered by  strong ocean currents, leading to unexpected movement or severe abort. To address this issue, the GENIoS\_Python software has been developed as a real-time autonomous glider navigation system. By employing flow predictions to generate waypoints, the software improves navigation accuracy and relieves heavy intervention of  glider pilots. This software displays multiple advantageous features, including close monitoring of glider status, customizable experiment settings, light computing requirements, stable communication with dockservers, extended operation time, and quantitative flow estimate comparisons. Future work will focus on data visualization to better interact with human pilots like plotting offset between the expected surfacing position and the actual one in real time.   We will also pay attention to incorporating more ocean flow models such as HFRadar and WERA. 

\bibliography{IEEEabrv, reference.bib}
\bibliographystyle{IEEEtran}

\end{document}